\documentclass[letterpaper, 10 pt, conference]{ieeeconf}  

\IEEEoverridecommandlockouts                              

\usepackage{todonotes}
\usepackage{graphics} 
\usepackage{epsfig} 
\usepackage{amsmath} 
\usepackage{amssymb}  

\title{\LARGE \bf
Deep Person Re-identification for Probabilistic Data Association in Multiple Pedestrian Tracking
}

\author{Brian H. Wang$^{1}$, Yan Wang$^{2}$, Kilian Q. Weinberger$^{2}$, and Mark Campbell$^{1}$
\thanks{\small This work was supported by the Office of Naval Research, under grant N00014-17-1-2175.}
\thanks{\small$^{1}$Autonomous Systems Lab, Department of Mechanical and Aerospace Engineering,
        Cornell University, Ithaca, NY 14850, USA.
        {\small \{bhw45, mc288\}@cornell.edu}}%
\thanks{\small$^{2}$Department of Computer Science,
        Cornell University, Ithaca, NY 14850, USA.
        {\small \{yw763, kqw4\}@cornell.edu}}%
}

\begin{document}

\maketitle
\thispagestyle{empty}
\pagestyle{empty}


\begin{abstract}

We present a data association method for vision-based multiple pedestrian tracking, using deep convolutional features to distinguish between different people based on their appearances. These re-identification (re-ID) features are learned such that they are invariant to transformations such as rotation, translation, and changes in the background, allowing consistent identification of a pedestrian moving through a scene. We incorporate re-ID features into a general data association likelihood model for multiple person tracking, experimentally validate this model by using it to perform tracking in two evaluation video sequences, and examine the performance improvements gained as compared to several baseline approaches. Our results demonstrate that using deep person re-ID for data association greatly improves tracking robustness to challenges such as occlusions and path crossings. 
\end{abstract}

\section{Introduction}




Visually tracking the motion of people through a scene over time is a critical capability for many applications involving camera-equipped robots or sensor networks. Examples range from an autonomous car tracking nearby pedestrians, to a team of aerial robots searching for moving people in a search-and-rescue mission. This problem can be broken down into two general stages: people must be detected in video frames, and these detections must then be linked together and used to estimate tracks over time. The first of these two tasks has been extensively studied with the usage of deep convolutional neural networks for object detection \cite{Girshick2014,Krizhevsky2012,Dai2016,Ren2015}. State of the art algorithms such as Mask-RCNN \cite{He2017} are capable of detecting people at the pixel level with near-human level accuracy, and have been found to generalize well to different scenes.

However, for multiple pedestrian tracking applications, it is not enough to just detect the presence of people - it is equally important to distinguish between individuals and correctly associate detections with currently tracked people. A robot which confuses individuals with one another could assume an incorrect number of people in its environment in an autonomous navigation or search-and-rescue scenario, or an intelligent sensor network could lose track of a person of interest in a security task. 

\begin{figure}
  \includegraphics[width=\columnwidth]{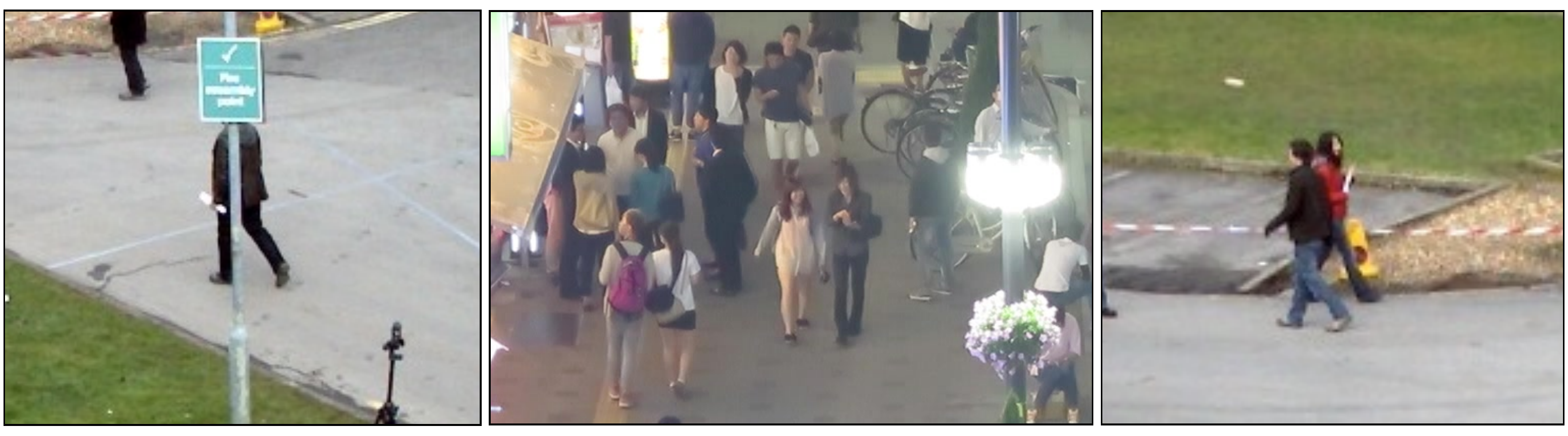}
  \caption{Examples of challenging situations for a multiple pedestrian tracker: a person being occluded by an obstacle, a large crowd, and two people crossing paths.
}\label{fig:tracking_challenges}
\end{figure}
This problem becomes highly challenging in the presence of occlusions, crowds, and pedestrians crossing paths with one another. See Figure \ref{fig:tracking_challenges} for an example of some of the situations that complicate multiple-person tracking. A method for modeling appearance and distinguishing between different people visually could provide key information to robustly handle these challenges.

Person re-identification (re-ID) provides a promising solution to this problem \cite{Zheng2015,Cheng2016,Li2017,Wang2018}. Given two images that each contain a person, a re-ID system can generate a likelihood that the two images are of the same person. Similar to people detection in a scene, re-ID is extremely intuitive for humans - consider how quickly most people can spot a friend in a crowd, or pick out a cameo appearance by a favorite celebrity in a movie. However, re-ID is challenging for a computer, and is therefore an active area of research in machine learning and computer vision. State of the art approaches to person re-ID involve modeling the appearances of individuals in a low-dimensional feature space, that is learned with a deep convolutional neural network trained on a large dataset of images of many different people. The network is trained explicitly so that images of the same person are mapped to close-by locations in feature space, while images of different people are mapped to far apart locations.

\begin{figure}
  \includegraphics[width=\columnwidth]{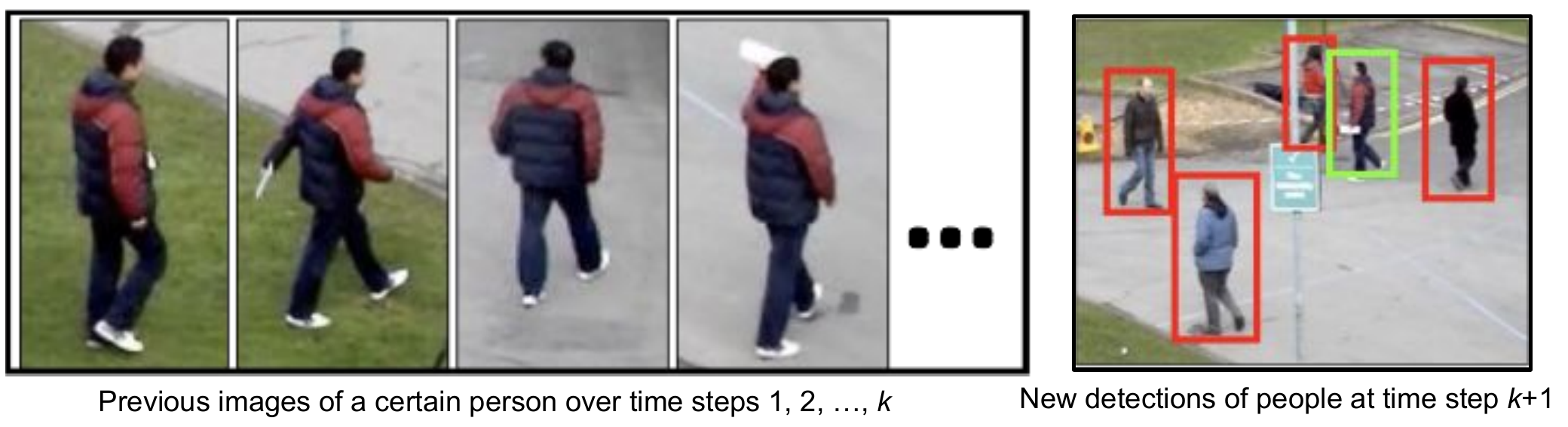}
  \caption{Given a set of previous detections of a person, deep person re-ID could hypothetically be used to re-identify this person within the set of detections seen in a new video frame.}\label{fig:reid_data_assoc_example}
\end{figure}

Modern person re-ID methods are becoming very successful as measured by performance on publicly available re-ID benchmark datasets \cite{Zheng2015,Zheng2016a,Ahmed2015,Cheng2016,Li2017,Wang2018}. However, the usefulness of deep re-ID methods for application in probabilistic tracking algorithms has not yet been extensively studied. Traditionally in pedestrian tracking, position information is used to association a new observation to a nearby tracked person. It is proposed here to additionally use appearance, via re-ID, as a cue to aid data association decisions, as seen in the example in Figure \ref{fig:reid_data_assoc_example}. This paper studies how well deep learning-based person re-identification improves data association in probabilistic multiple-person tracking.

\section{Related Work}



\subsection{Probabilistic Data Association}

Data association is a difficult problem in multiple-object tracking, and several proven approaches exist that achieve respectable performance. The Rao-Blackwellized particle filter (RBPF) probabilistically evaluates multiple data association decisions at each time step, and propagates multiple hypotheses of assignments forward in time \cite{Sarkka,Schulz2003,Miller2011a}. Multiple Hypothesis Tracking (MHT) works similarly, but rather than performing probabilistic sampling, it grows a tree of possible hypotheses according to deterministic branching decisions \cite{Blackman2004a,Kim2015}.

Some tracking methods perform data association over long periods of time, using probabilistic graphical models where more nodes are added to the graph as the length of time considered increases \cite{Zhang2008,Yang2017,Keuper2016}. These methods are able to jointly reason about multiple objects in the scene over multiple time steps, and therefore generally achieve higher overall accuracy compared to online methods. However, such methods are not well-suited for applications in robotics or intelligent sensor networks, which generally require tracking to be performed recursively one frame at a time, as the system captures video data sequentially.

\subsection{Person Re-identification}

\begin{figure}
  \includegraphics[width=\columnwidth]{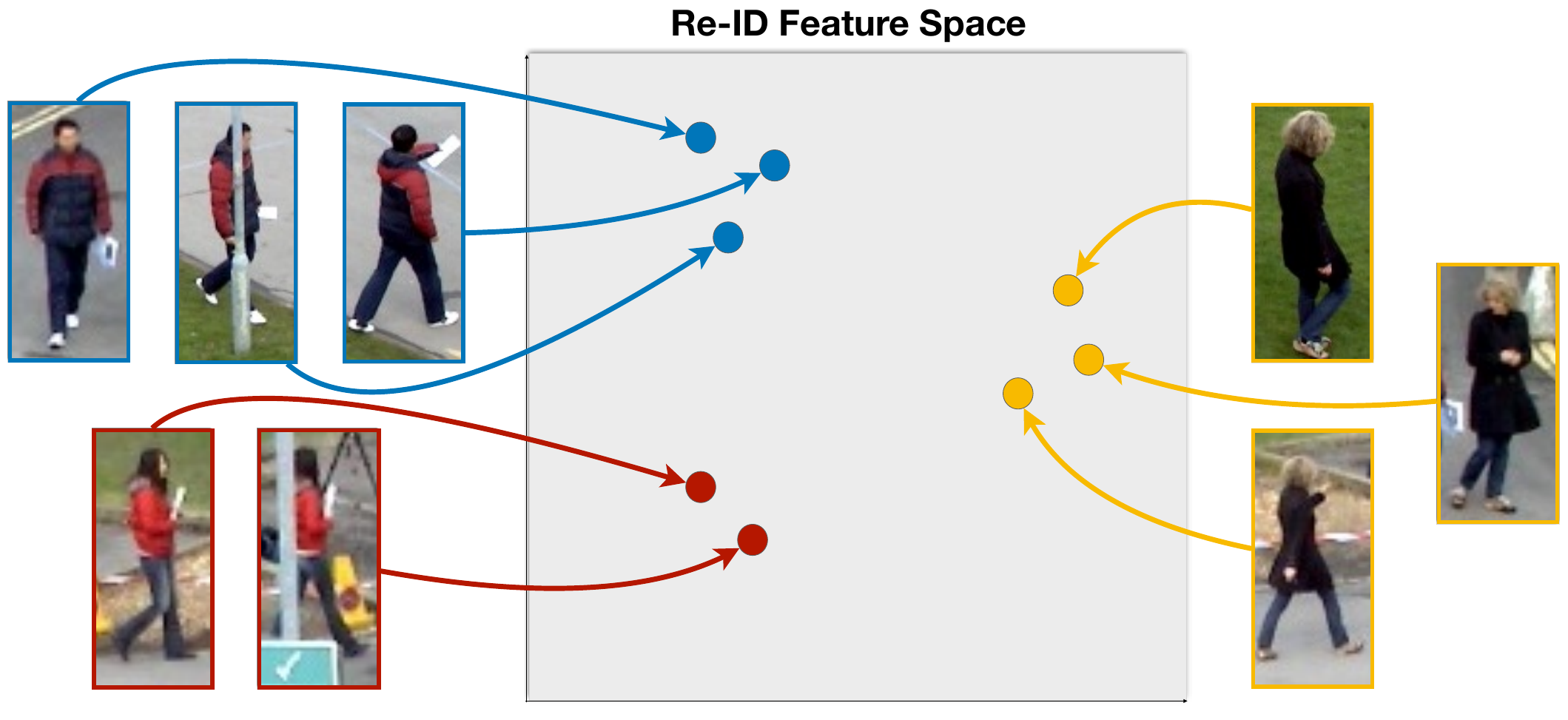}
  \caption{Visualization of how images of the same person will map to nearby points in re-ID feature space. In this figure, the re-ID feature space is shown flattened down to two dimensions.}\label{fig:reid_example}
\end{figure}

The general goal of machine learning-based person re-identification is to learn a method for mapping images of people to low-dimensional feature vectors. These feature vectors should have the property that two images which are of the same person map to nearby feature points (as measured by Euclidean distance), while images of two different people map to feature points that are far apart. This is illustrated in Figure \ref{fig:reid_example}. Note that ideally, re-ID feature representations should be robust to changes in the background, pose, or orientation relative to the camera, as well as to partial occlusions by obstacles.

People are able to perform re-identification by looking at visual cues such as clothing color, facial features, body shape, and distinctive accessories, to name just a few. Deep convolutional networks have been shown \cite{Girshick2014} to learn and extract high quality features from natural images, which are invariant to small local transformations such as rotation and translation. For our re-ID task, these invariants are crucial, as they help ensure that a person's identity remains unchanged as he or she moves across a scene. Indeed, many successful approaches to re-ID have used deep convolutional neural networks to map images to re-ID feature vectors \cite{Ahmed2015,Cheng2016,Li2014,Li2017}. Researchers have also created a number of high-quality benchmark datasets, for the purpose of training and evaluating these methods \cite{Ristani2016,Zheng2015,Zheng2016a}.

\subsection{Deep Learning for Multiple Pedestrian Tracking}
Deep learning has also been applied directly to the problem of multiple pedestrian tracking. In particular, recurrent neural networks (RNNs) have been employed for tracking \cite{Milan2016}, due to their ability to effectively process time series data. Sadeghian et al. \cite{Sadeghian2017} use RNNs to learn motion, appearance, and interaction-based cues that indicate similarity between new observations and previously tracked pedestrians. While shown to be effective given sufficient training data, it remains to be seen whether data-driven approaches to data association and tracking are viable for robotic applications, which typically involve significant variations in environments and operating conditions, necessitating an exponentially greater amount of training data for successful performance.




\section{Re-ID-based Data Association}


\subsection{Probabilistic Data Association}

\begin{figure*}
  \includegraphics[width=\textwidth]{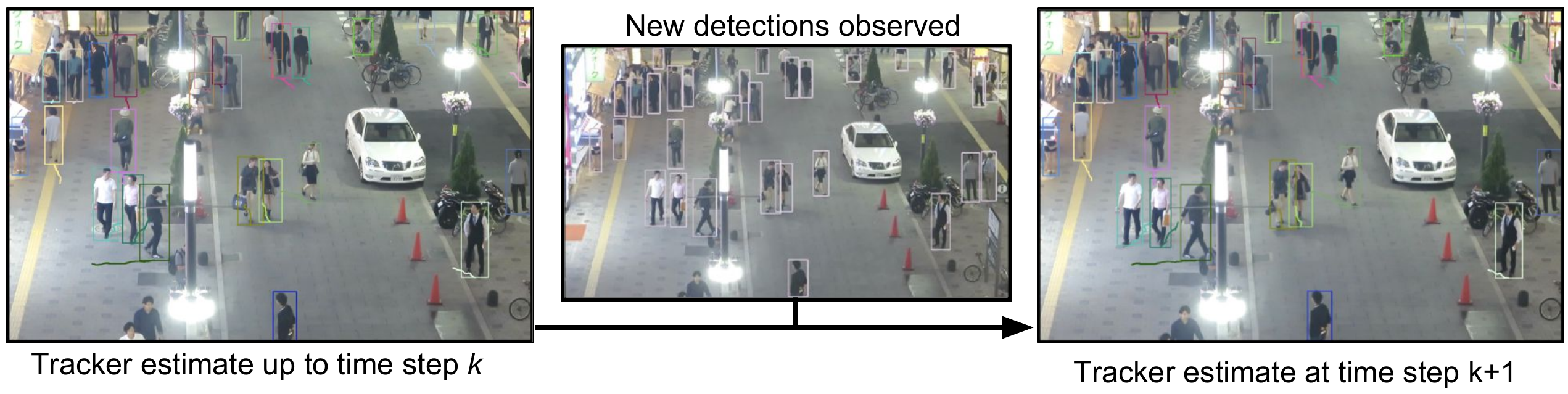}
  \caption{Example of the data association problem for vision-based multiple pedestrian tracking. The left image shows estimated tracks of pedestrians up to the current time step. When a new set of detections is observed (center), each new detection must be either associated with a previously tracked person or used to initialize a new track, thereby updating the track estimates by one time step (right).}\label{fig:data_assoc_example}
\end{figure*}

In this section, we begin by defining the general data association problem within the context of multiple object tracking. Given a set of measurements $Z(k)$ and a set of tracked objects $X(k)$, where $k$ is the current time step, the goal of data association is to determine which measurement was generated by which tracked object. Individual measurements are denoted by $z_i(k)\in Z(k)$, $i=1,\dots,m_k$, and estimated object tracks are denoted by $x_j(k)\in X(k)$, $j=1,\dots,n_k$.

For the pedestrian tracking task, each time step $k$ corresponds to a distinct video frame, and the measurements $Z(k)$ are a set of bounding boxes or segmentation masks found by a computer vision person detector. Data association is then a discrete assignment problem, where each detection must be assigned to a tracked person.  
\def\asgn{\theta_{ij}}
\def\asgnprob{P\left(\asgn(k) \vert z_i\left(k\right),x_j\left(k\right) \right)}

This problem is shown more specifically in Figure \ref{fig:data_assoc_example}. Each member of the set of new detections (Fig. \ref{fig:data_assoc_example} center) must be either assigned to a previous track (Fig. \ref{fig:data_assoc_example} left). In order to perform data association, a likelihood value is typically defined for each possible assignment from measurement $z_i(k)$ to track $x_j(k)$, denoted as $a_{ij}(k)$. Let $\asgn(k)$ be the event that the detection $z_i(k)$ was generated by person $x_j(k)$. The likelihood values $a_{ij}(k)$ are then defined as
\begin{equation}\label{eq:assoc_likelihoods}
  a_{ij}(k) \equiv \asgnprob
\end{equation}
for $i=1,\dots,m_k$ and $j=1,\dots,n_k$. Note that a separate likelihood model should also be defined for the event where detection $i$ is used to initialize a new track, as this falls outside the data association problem.

Equation (\ref{eq:assoc_likelihoods}) defines the probability of an assignment event only conditioned on the current detection $z_i(k)$ and track state $x_j(k)$. This is the approach used in nearest-neighbors and one-shot data association strategies \cite{BarShalom2009}. Other approaches, such as the Rao-Blackwellized particle filter and MHT, operate over the full history of measurements, $Z(1),\dots,Z(k)$, using (\ref{eq:assoc_likelihoods}) only for assignments at the last time step. Typically, even these approaches define recursive solutions, which closely examine the most recent event in (\ref{eq:assoc_likelihoods}). In this work, we focus on the individual likelihood shown in (\ref{eq:assoc_likelihoods}), but importantly, this approach can be generalized to any data association method \cite{BarShalom2009,Sarkka,Miller2011a,Blackman2004a} 

Since one of the events $\theta_{ij}(k)$ must explain the source of detection $z_i(k)$, and a detection cannot come from two different people, these events are mutually exclusive and exhaustive, and

\begin{equation}
\sum_{j=1}^{n_k} a_{ij}(k) = 1 \quad \forall i=1,\dots,m_k.
\end{equation}

The association likelihoods $a_{ij}$ then depend on the selection of the likelihood model $\asgnprob$ in (\ref{eq:assoc_likelihoods}. Since many of the sensors traditionally used for robotics and tracking applications provide location measurements, data association likelihoods typically depend only on the location of the sensor measurement relative to the expected object position. As an example, points returned by a lidar sensor can only be reliably associated with tracked objects according to the point positions \cite{Miller2007}. In cases such as these, the sensor likelihood model can be accurately approximated as a Gaussian centered on the expected object position.

When using object detections in a video frame, the measurement $z_i(k)$ includes detection position as well as size (for a bounding box) or shape (for a segmenting mask). However, these detections also contain useful information about the appearance of detected people. Thus, camera detection measurements can be decomposed into two independent measurements: a position component $z^{pos}_{i}(k)$ and an appearance component $z^{app}_{i}(k)$. The measurement likelihood model then becomes
\begin{equation}\label{eq:p_asgn_given_z}
  a_{ij}(k) = P\left(\theta_{ij}(k)\vert z^{pos}_i(k), z^{app}_i(k)\right).
\end{equation}
Using Bayes rule, (\ref{eq:p_asgn_given_z}) can be rewritten as
\begin{equation}
   a_{ij}(k) = \alpha P\left(z^{pos}_i(k), z^{app}_i(k)\vert \theta_{ij}(k)\right) \times P\left(z^{pos}_i(k),z^{app}_i(k)\right),
\end{equation}
where $\alpha$ is a normalization factor. The term $P\left(z^{pos}_i(k),z^{app}_i(k)\right)$ represents any prior information that may be available about the likelihood of observing detections; it is common to simply use a uniform distribution over all possible assignments \cite{Miller2011a}.
Assuming a uniform prior, and noting that the position and appearance components of the measurement are independent of one another, the assignment likelihood values are therefore
\def\poslhood{P\left(z^{pos}_i(k)\vert \theta_{ij}(k)\right)}
\def\applhood{P\left(z^{app}_i(k)\vert \theta_{ij}(k)\right)}
\begin{equation}\label{eq:assignment_model_pos_app}
   a_{ij}(k) = \alpha \poslhood\applhood.
\end{equation}
The likelihood of detection $z_i(k)$ being observed at a certain position $z^{pos}_i(k)$ given that it has been assigned to person $j$ can be accurately modeled as Gaussian, with the likelihood decreasing as distance to the estimated position of person $j$ increases.

Deriving the detection appearance model $\applhood$ requires augmenting the estimated state vector for each tracked person to include information on the person's appearance, in the form of re-ID features. The augmented feature vector for person $j$ is defined as
\def\xaug{\widetilde{x}}
\begin{equation}
  \xaug_j(k) = \begin{bmatrix}
    x_j(k)\\
    f_j(k)
  \end{bmatrix},
\end{equation}
where $f_j(k)$ is a vector in re-ID feature space. For tracking using bounding box detections, the positional component $x_j(k)$ stores the tracked person's estimated position, velocity, and bounding box size. In our data association approach, the stored feature vector $f_j(k)$ is formed from a moving average of all feature vectors from detections assigned to person $j$. 

\subsection{Re-ID Likelihood Model}

One final question in the data association likelihood model is how to transform Euclidean distances between re-ID feature points into the likelihood values $\applhood$ used in (\ref{eq:assignment_model_pos_app}). When new detections are received, they are each converted to re-ID feature vectors using a deep re-ID model. Each tracked person has an averaged reference feature vector $f_j(k)$ associated with them. 

Then, given a new detection $z^{app}_i(k)$, which is converted into a re-ID feature vector $g_{i}(k)$, the likelihood of assigning it to person $j$ can be calculated using the softmin function over all tracked person reference vectors, giving
\begin{equation}
  P\left(\theta_{ij}(k)\vert z^{app}_i(k)\right) = \beta_i\exp\left(-\Vert g_i(k) - f_j(k)\Vert\right).
\end{equation}
$\beta_i$ is a normalization term used to ensure that $\sum_{j=1}^{n_k}P\left(\theta_{ij}(k)\vert z^{app}_i(k)\right)=1$, and therefore
\begin{equation}
  \beta_i = \left(\sum_{j=1}^{n(k)} \exp\left(-\Vert g_i(k) - f_j(k)\Vert\right)\right)^{-1}.
\end{equation}
The softmax function is commonly used in machine learning to convert from feature vectors to class likelihoods in multiclass classification problems; since we are here interested in the minimal distance between pairs of feature vectors, we instead use a softmin to form a discrete probability distribution for $\applhood$ over the $n_k$ possible person assignments. The full data association likelihoods shown in (\ref{eq:assignment_model_pos_app}) are then finally computed by multiplying these appearance similarity-based probabilities together with the Gaussian detection position likelihoods.

\subsection{Deep Anytime Re-ID (DaRe)}

In our multiple pedestrian tracker, we use the Deep Anytime Re-ID (DaRe) architecture from Wang et al. \cite{Wang2018} to perform person re-identification through transforming images of detected people to re-ID feature vectors. In addition to achieving state of the art performance on re-ID benchmarks, DaRe is particularly well-suited for robotic applications, because it utilizes varying amounts of computational resources depending on the person being re-identified. As an example, a person who is wearing a distinctive outfit and is clearly visible is identified using only the first stage of the DaRe convolutional neural network; someone else who is partially occluded, blurred by camera motion, or wearing clothes that blend into the background is identified using a greater number of convolutional layers.

At each stage, DaRe calculates a confidence value for the re-identification result, and stops computation when sufficiently confident. The overall effect is a significant reduction in computation time, as compared to approaches that apply computation uniformly to each person. Our implemented model is trained on the MARS dataset \cite{Zheng2016a}, and makes uses of a dense convolutional neural network (DenseNet) architecture \cite{Huang2016}.

\section{Experimental Validation}


In order to experimentally examine the effects of incorporating appearance re-ID into a data association likelihood model, we present results from implementing the above described data association strategy within a Rao-Blackwellized particle filter. The algorithm tracks multiple moving pedestrians within two video sequences showing complex street scenes. In order to understand the impact of re-ID on data association, results are presented using four data association methods: detection position only, deep re-ID likelihood only, position along with a simple appearance model, and finally position combined with deep re-ID. Results are analyzed quantitatively using various numerical metrics of tracking performance, and qualitatively by examining specific cases where the inclusion of deep re-ID improves tracker robustness to certain difficulties.

\subsection{Rao-Blackwellized Particle Filter}

There are many different algorithms for performing data association and multiple-object tracking, based on the association likelihoods defined in (\ref{eq:assignment_model_pos_app}). In order to consider uncertain data association decisions, our experiments utilize a Rao-Blackwellized particle filter (RBPF) \cite{Miller2007,Miller2011a}, which samples data association hypotheses (in the form of particles) based on the likelihood model in (\ref{eq:assignment_model_pos_app}). Individual objects are tracked using efficient parametric trackers, and decisions on when to initiate new tracks from detections are also made probabilistically according to the detection association likelihoods. Detailed treatments of the RBPF can be found in S{\"a}rkk{\"a} et al. \cite{Sarkka} and Miller and Campbell \cite{Miller2007}.

Our RBPF uses separate linear motion model Kalman filters to track each individual person. Linear motion is only an approximation for the true patterns of pedestrian motion; person detectors provide frequent enough measurement updates to correct for model inaccuracy and adequately track pedestrians. The RBPF could certainly be extended with a more complex physics-based or data-driven motion model; this is outside the focus of this paper.

\begin{figure}
  \includegraphics[width=\columnwidth]{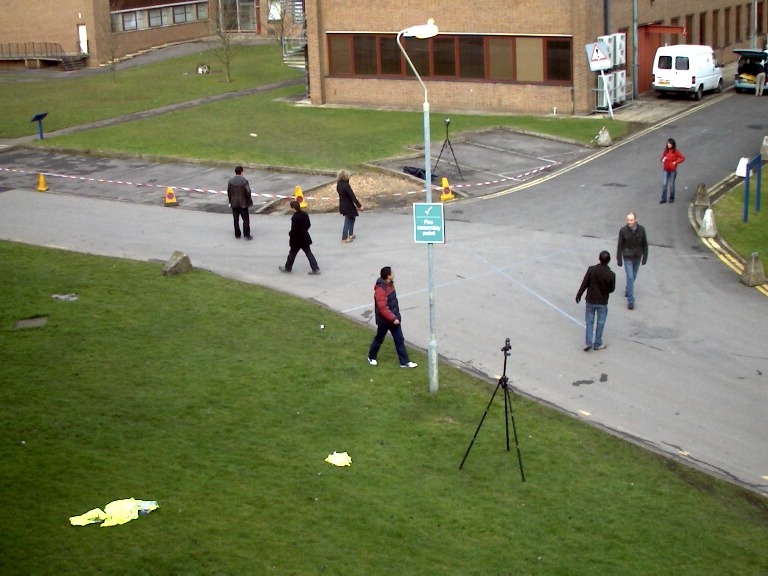}
  \caption{A frame from the PETS09-S2L1 video sequence.}\label{fig:pets780}
\end{figure}

\begin{figure}
  \includegraphics[width=\columnwidth]{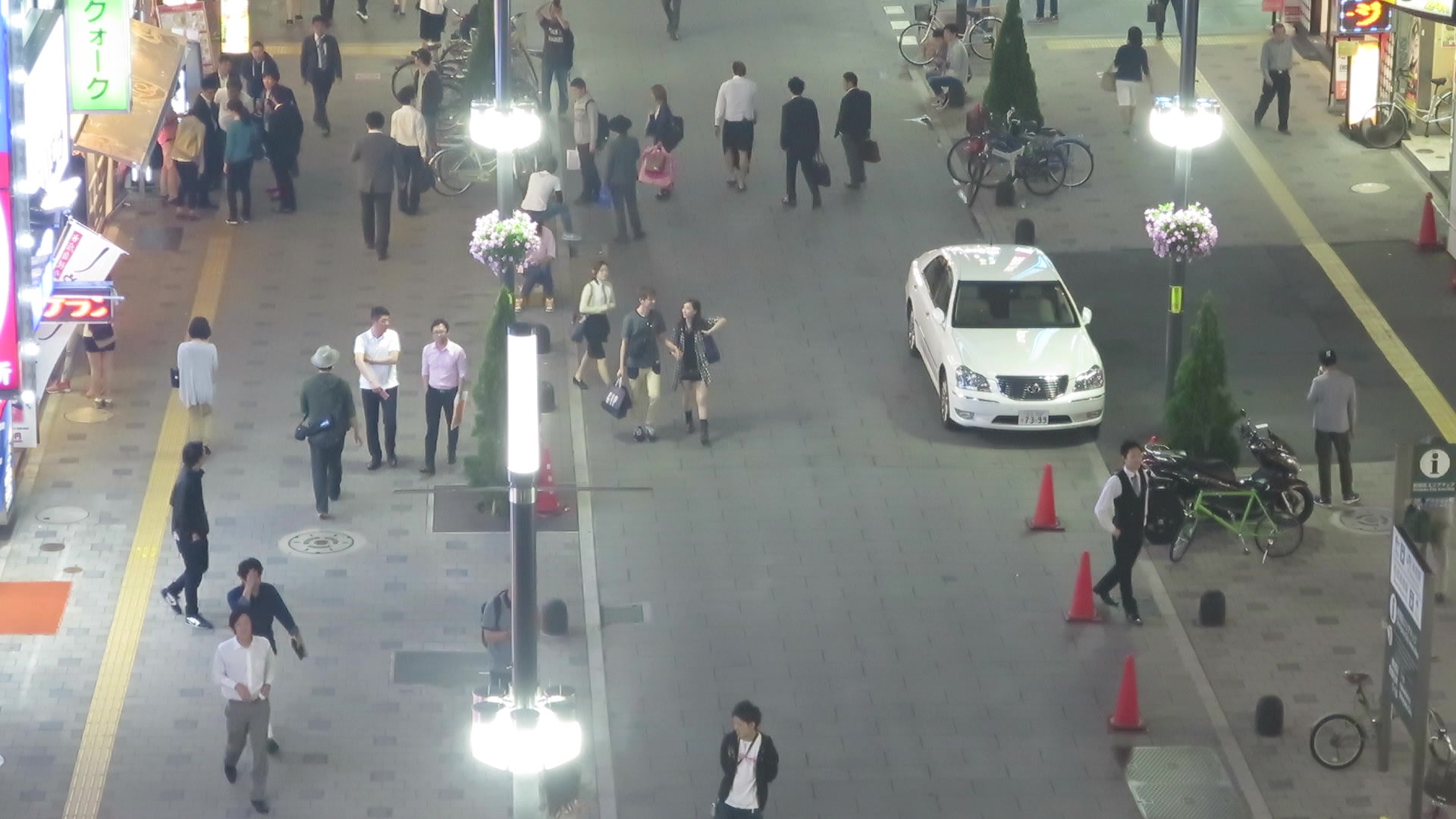}
  \caption{A frame from the MOT17-04 video sequence.}\label{fig:mot1704}
\end{figure}

In order to demonstrate the applicability of deep person re-ID to multiple-pedestrian tracking, the RBPF tracker was applied to a pair of challenging video sequences taken from the Multiple Object Tracking Challenge (MOTC) benchmark \cite{Leal-Taixe2015,Milan2016a}. The PETS09-S2L1 and MOT17-04 sequences are used, shown in Figures \ref{fig:pets780} and \ref{fig:mot1704} respectively. These sequences show people walking in various patterns, with occlusions by each other and by other objects, and therefore provide a useful evaluation for tracking. Detection bounding boxes and ground truth person locations are provided for each sequence, with the ground truth data being used only for performance evaluation after running the tracker.

It should also be noted that our deep re-ID architecture is trained only on the MARS dataset \cite{Zheng2016a}, and not on images of people from either of the MOTC videos. Therefore, this experiment also studies the ability of our deep re-ID system to generalize to new images of previously unseen people.

For each sequence, several data association strategies are compared in a controlled study. The data association variants we use are as follows: Using position only, using deep re-ID only, using a benchmark color histogram appearance model, and finally, using position along with deep re-ID. The benchmark appearance model uses a color histogram in HSV color space to judge visual similarity between tracked people and new detections, and is included in order to evaluate the performance afforded by deep re-ID as compared to a much simpler, non-machine learning-based model.

\begin{figure}
  \includegraphics[width=\columnwidth]{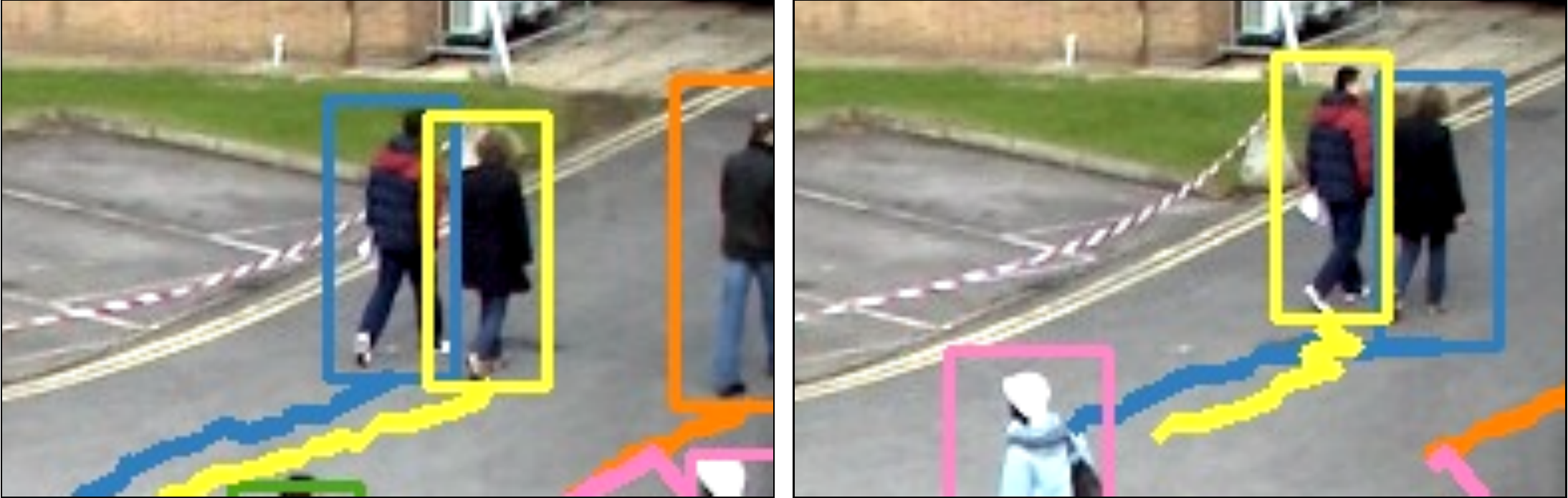}
  \caption{An example of an ID switch caused by an error from position-only data association. Colors represent different tracks. As two people walk side-by-side, the position-only RBPF tracker initially tracks them correctly (left), but later on switches their track IDs (right).}\label{fig:idswitch}
\end{figure}

Performance is measured quantitatively with the widely used CLEAR MOT metrics, consisting of multiple object tracking accuracy (MOTA) and multiple object tracking precision (MOTP) \cite{Bernardin2008}. MOTA decreases as the rate of false positives, false negatives, or ID switches increase. An example of an ID switch is shown in Figure \ref{fig:idswitch}, where the tracker mistakenly swaps the tracks of two different people. The hypothesis behind using deep re-ID for data association is that it should increase MOTA by decreasing the number of ID switches in particular, since ID switches are tracker errors that are directly caused by data association mistakes. In cases such as that seen in Figure \ref{fig:idswitch}, where two people walk next to one another, data association based on detection positions only is extremely ambiguous. A small amount of detection noise or unexpected motion could easily cause an ID switch. However, incorporating re-ID into data association should intuitively protect against such occurrences, as the tracker could then use appearance information to disambiguate nearby people from one another.

MOTP measures the precision with which people's exact locations are known. This is primarily influenced by the precision of the object detector; a detector that fits bounding boxes more closely around people, or avoids bounding boxes entirely in favor of segmenting masks, would attain higher precision. On the other hand, the choice of data association strategy has no direct influence on precision, and so we do not expect a large effect on MOTP. 



\newcommand\B\bfseries 
\begin{table}
\caption{Results on sequence PETS09-S2L1}
\label{table:seq_pets}
\begin{tabular}{l|l l l l l l}
\hline
 Method & MOTA & MOTP & FP & FN & ID Sw.\\
\hline
 Pos. only        & 0.905 & 0.644 & 154 & 260 & 30 \\
 Re-ID only     & 0.109 & 0.544 & 597 & 3434 & 112 \\
 Pos.+Hist. & -0.168 & 0.560 & 4801 & 411 & 218 \\ 
 \B Pos.+re-ID & \B 0.929 & \B 0.656 & \B 114 & \B 210 & \B 6 \\
\hline
\end{tabular}
\end{table}

\begin{table}
\caption{Results on sequence MOT17-04}
\label{table:seq_1704}
\begin{tabular}{l|l l l l l l}
\hline
  Method & MOTA & MOTP & FP & FN & ID Sw.\\
\hline
 Pos. only  & 0.445 & 0.820 & 7831 & 18390 & 160 \\
 Re-ID only & 0.156 & 0.793 & 2919 & 36684 & 535 \\
 Pos.+Hist. & 0.195 & 0.738 & 21977 & 15437 & 862 \\ 
 \B Pos.+Re-ID  & \B 0.533 & \B 0.863 & \B 2253 & \B 19871 & \B 83 \\
\hline
\end{tabular}
\end{table}


\subsection{Quantitative Analysis}

Tracking results from running the RBPF tracker on the two evaluation sequences are shown in Tables \ref{table:seq_pets} and \ref{table:seq_1704}. MOTA, MOTP, FP, FN, and ID Sw. indicate Multiple Object Tracking Accuracy, Multiple Object Tracking Precision, false positives, false negatives, and ID switches, respectively. The PETS09-S2L1 sequence includes 795 video frames showing 19 pedestrians walking, with a total of 4650 ground truth person annotations over all frames. This video sequence is fairly sparse; at any given time, 2 to 8 pedestrians are seen in the video frame simultaneously. The MOT17-04 sequence includes 1050 frames showing 83 pedestrians, with 47557 total ground truth annotations. This sequence is much more complex than the PETS09-S2L1 video; as seen in Figure \ref{fig:mot1704}, at times upwards of 30 people can be observed in the video frame, greatly increasing the difficulty of tracking.

The most significant effect of combining re-ID with position-based data association is a reduction in the number of ID switches, validating our earlier hypothesis. Compared to data association with position only, using position as well as re-ID caused a drop in ID switches from 30 to 6 in the PETS09-S2L1 sequence (an 80\% reduction), and from 160 to 83 in the MOT17-04 sequence (a 48\% reduction).

In order to realize the benefits of re-ID data association, it is not enough to use re-ID without position information, or to use a simple color histogram appearance model. Both approaches perform poorly. In the PETS09-S2L1 sequence, the color histogram method creates so many false tracks that its false positive count exceeds the total amount of ground truth annotations, causing negative MOTA overall. The poor performance of this method could be explained by the fact that color histograms are heavily influenced by colors in the background, as well as by the color of large objects such as a person's coat. In contrast, a re-ID feature representation can be learned such that it ignores irrelevant information. 

As expected, MOTP is not significantly affected by the inclusion of deep re-ID for data association as opposed to the position-only setting, though we do observe a modest increase. MOTP does however drop when MOTA is very low, as in the re-ID only and color histogram data association cases.
\subsection{Qualitative Analysis}

\begin{figure}
  \includegraphics[width=\columnwidth]{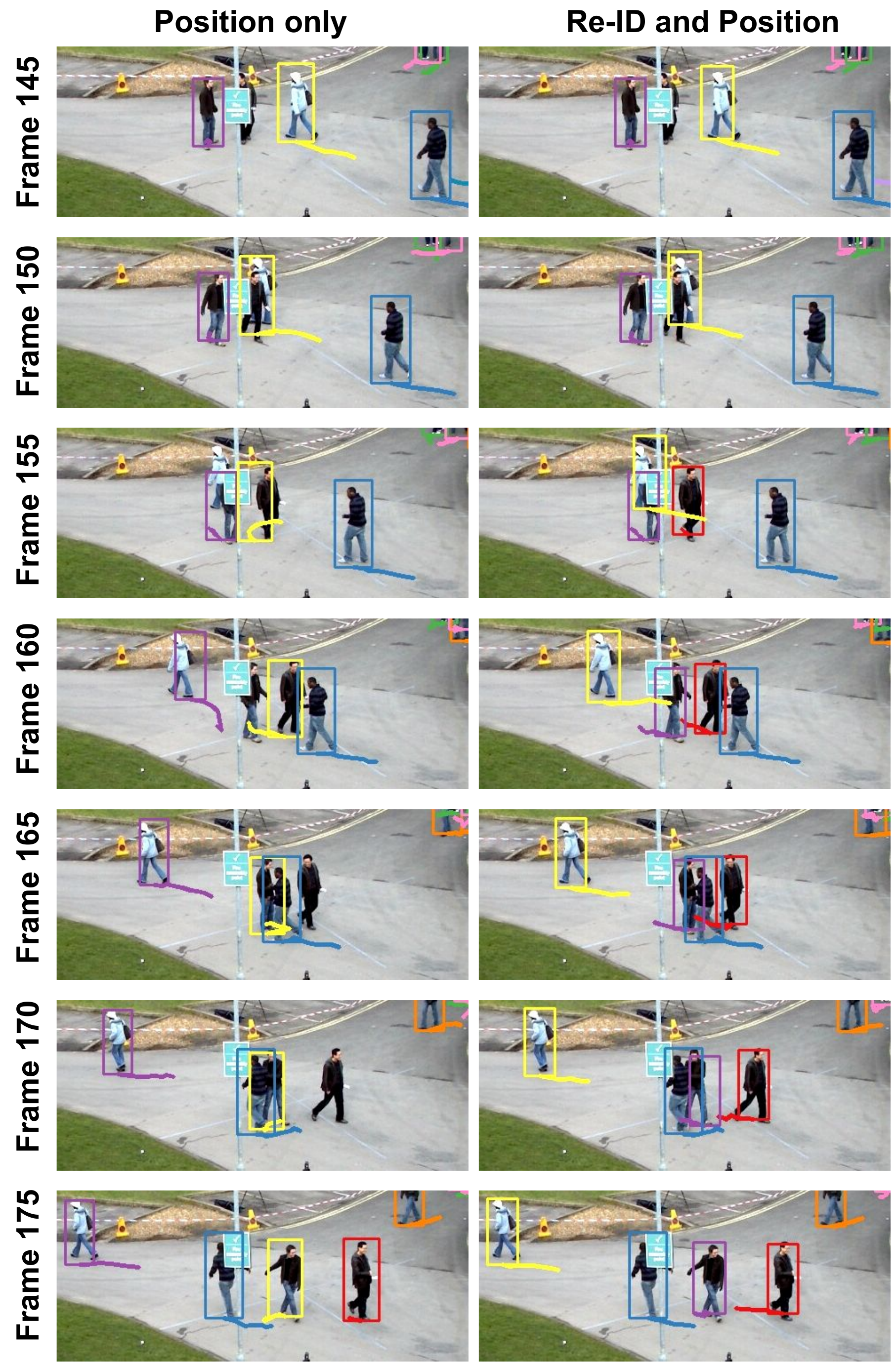}
  \caption{Frames 145 to 175 of the PETS09-S2L1 sequence, showing tracking results with position-only data association (left) and position with re-ID data association (right). Frames are cropped from the originals to better focus on this group of people.}\label{fig:comparison_pets}
\end{figure}

We next present a qualitative analysis of a scenario from the tracking results, in order to illustrate how deep re-ID assists data association and reduces the occurrence of ID switches, as seen from our quantitative results. Figure \ref{fig:comparison_pets} shows a series of frames from the PETS09-S2L1 sequence, where a group of people walk across the scene. Tracking results with and without re-ID data association are shown, with the left-side column showing position-only data association tracking, and the right-side column showing tracking using position and re-ID. 

In frame 150 of the position-only case, an ID switch occurs as one person (in the yellow bounding box) walks behind a signpost while another person, who previously had been lost by the tracker due to an extended period of occlusion, walks out from behind it at nearly the same location. The yellow box can be seen to have switched from one person to the other in frame 155. As the first person walks out from the other side of the signpost in frame 160, a second ID switch from a different person occurs, again due to occlusion, as seen from the movement of the purple box.  In frame 165, the yellow box again switches between people, causing one person to be missed by the tracker in frame 170 until a new track is initialized in frame 175. When re-ID as well as position are used for data association, these ID switches do not occur, and the tracker is able to successfully use the distinct appearances of these people to perform reliable tracking even in the presence of occlusions and path crossings. 

\section{Conclusion}

A general approach to augmenting traditional sensor likelihood models with deep person re-identification is presented, for application in multiple person tracking. We describe the process of converting images of people into convolutional feature vectors, using a learned deep re-ID model, and show how these feature vectors can be used within the general data association framework. Our results indicate that person re-ID significantly increases tracking performance as compared to data association that uses detection position only, according to quantitative measures of tracking accuracy and consistency. In particular, the usage of deep re-ID for data association is seen to cause an 80\% drop in ID switches for tracking in the PETS09-S2L1 video sequence, as well as a 48\% reduction in the much more crowded and complex MOT17-04 sequence. The usage of deep re-ID is additionally seen qualitatively to increase tracking robustness to difficulties such as occlusions and path crossings.




\addtolength{\textheight}{-9cm}   







\bibliographystyle{IEEEtranS}
\bibliography{references}

\end{document}